\documentclass[conference]{IEEEtran}
\usepackage{times}
\usepackage{booktabs,caption}
\usepackage[flushleft]{threeparttable}
\usepackage{multirow}
\usepackage{makecell}
\usepackage{graphicx}

\usepackage[numbers]{natbib}
\usepackage{multicol}
\usepackage[bookmarks=true]{hyperref}

\begin{document}

\title{Singing the Body Electric: The Impact of \\Robot Embodiment on User Expectations}


\author{\authorblockN{Nathaniel Dennler}
\authorblockA{Computer Science Department\\
University of Southern California\\
Los Angeles, CA\\
Email: dennler@usc.edu}
\and
\authorblockN{Stefanos Nikolaidis}
\authorblockA{Computer Science Department\\
University of Southern California\\
Los Angeles, CA\\
Email: stefanosnikolaidis@gmail.com}
\and
\authorblockN{Maja Matari\`c}
\authorblockA{Computer Science Department\\
University of Southern California\\
Los Angeles, CA\\
Email: mataric@usc.edu}}


%

\maketitle

\begin{abstract}
Users develop mental models of robots to conceptualize what kind of interactions they can have with those robots. The conceptualizations are often formed before interactions with the robot and are based only on observing the robot's physical design. As a result, understanding conceptualizations formed from physical design is necessary to understand how users intend to interact with the robot. We propose to use multimodal features of robot embodiments to predict what kinds of expectations users will have about a given robot's social and physical capabilities. We show that using such features provides information about general mental models of the robots that generalize across socially interactive robots. We describe how these models can be incorporated into interaction design and physical design for researchers working with socially interactive robots.

\end{abstract}

\IEEEpeerreviewmaketitle

\section{Introduction}
Mental models of interactions with systems, including robots, are instrumental in allowing users to naturally interact with arbitrarily technically complex systems \cite{rueben2021mental}. Human-computer interaction has successfully used the concept of {\it design metaphors} to develop visual interfaces and interactions that are easy for users to learn \cite{voida2008re, jung2017metaphors, khadpe2020conceptual, kim2020conceptual}.

Understanding how users expect to interact with robots, compared to computers, is especially challenging because robots are physically embodied; they have a wider variety of form factors and interaction affordances than typical computer systems. While these additional modes of communication increase complexity, they additionally contribute to the increased social presence of robots compared to computers \cite{deng2019embodiment}. In this work, we propose to leverage information about how a robot is physically embodied to understand how people form mental models of that robot. This work leverages the Metaphors to Understand Functional and Social Anticipated Affordances (MUFaSAA) dataset \cite{dennler2023design} that contains 165 robot embodiments and their associated ratings of social and functional attributes to understand users' mental models. 

We contribute a set of models that can predict how people expect robots to behave socially and functionally, using image-based and text-based features. We show that using pre-trained image features can perform as well as using a hand-crafted feature set, reducing the labor required to annotate these features. We also provide insights for how this model can be used in robot design and interaction design.

\section{Background}
This section provides a brief overview of the concepts of embodiment in robotics and users' mental models of robots. 

\subsection{Robot Embodiment}
Robots are inherently different from computer-based agents because they are situated in the physical world and have the ability to interact with it, navigate in it, and/or manipulate it. Due to this interactive nature, robots have a stronger social presence and can leverage additional modes of communication inaccessible to other forms of technology, including proxemics, gaze, and gesture \cite{deng2019embodiment}. Previous work in both psychology and robotics has shown that people form expectations from initial observations of new technologies even before extensive use \cite{fiske2007universal,carpenter2009gender,moon1996real,kwon2016human}. The physical design of the robot, i.e., its embodiment, is a key component of how users form expectations of robots capabilities and possible interactions with those robots. Consequently, for robots to be effective, they must understand the social and functional expectations that users place on them so that they can meet those expectations appropriately. Failing to do so negatively impacts adoption of these systems \cite{cha2015perceived,davis1989perceived,kwon2016human}.

While previous works has examined ways to model how features of embodiment affect perceptions of particular axes of robot identity, such as age/gender \cite{perugia2022shape} and anthropomorphism \cite{phillips2018human}. Our work is the first to show how embodiment dictates both functional and social expectations simultaneously, and that these expectations can be understood from easy-to-generate features in addition to hand-crafted features. 

\begin{figure*}
    \vspace{-1.5cm}
    \centering
    \includegraphics[width=\linewidth]{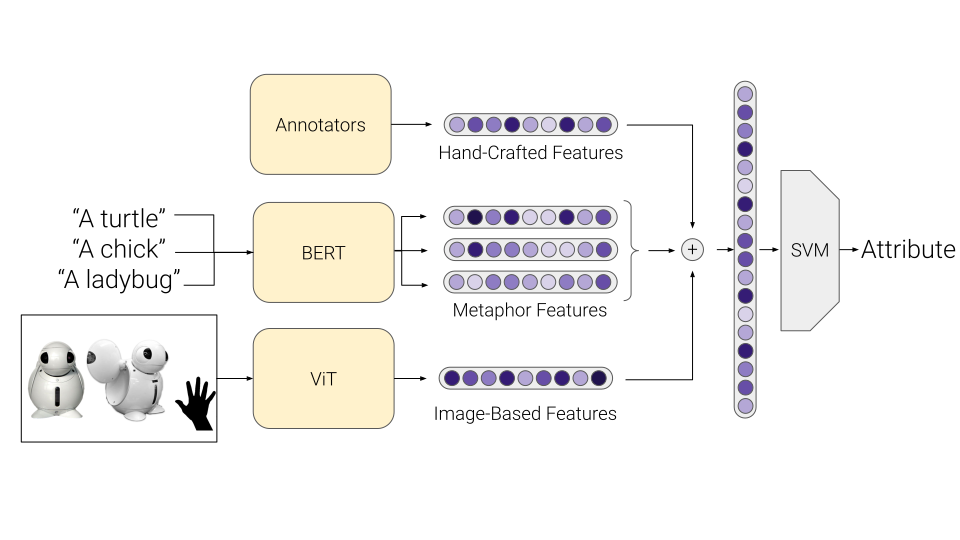}
    \vspace{-1.9cm}
    \caption{Process for generating features from the MUFaSAA dataset.}
    \label{fig:overview}
\end{figure*}

\subsection{Mental Models and Design Metaphors}

Mental models are conceptual frameworks that people automatically develop to understand how they can interact with other agents \cite{kiesler2002mental}. Previous work has shown that users with mental models that accurately represent the complex underlying system are more effective at using those systems \cite{kieras1984role}. Mental models are often based on capabilities robots are expected to perform \cite{rueben2021mental} and are formed before interaction, but are updated as users interact with systems and learn more about how systems work.  However, even after interaction users can still form and retain incorrect mental models of robots' real capabilities. For example, past work has shown that robots using speech are expected to perform better physical manipulation despite the fact that those two capabilities are technically unrelated  \cite{cha2015perceived}. Understanding users' initial mental models is important for calibrating robot capabilities, in order to avoid misleading user expectations.

Design metaphors are often used to set expectations of new technologies; they associate unfamiliar systems with familiar and related concepts to provide a user with schemas to interact with novel systems. For example, in one study, identical chatbots were described with different design metaphors (e.g., ``a toddler", ``a trained professional", ``an inexperienced teenager", etc.), shaping user perceptions of the chatbot's warmth and competence, thereby affecting both the users' pre-interaction intention to use the chatbot and their subsequent intention to adopt the chatbot post-interaction despite being the same technical implementation~\cite{khadpe2020conceptual}. By studying the metaphors people use to understand robot embodiment, we hope to gain insight into how people expect to interact with a given robot embodiment.

\section{Methods}

Our work leverages the MUFaSAA dataset to predict the psychological constructs that describe the social and functional expectations users place on robots, representing the users' mental models of the robot. We outline our process for predicting these expectations below.

\subsection{MUFaSAA Dataset Description}

The MUFaSAA dataset is a collection of 165 socially interactive robots \cite{dennler2023design} that have been developed for research or as consumer products. All robots have a standardized image representation that includes a front and side view with a height reference, a set of hand-coded design features (see \cite{dennler2023design} for feature descriptions, annotator information, and interrater reliability statistics), and a set of three design metaphors that participants used to describe the robot. Each robot also contains ratings of the three constructs from the validated Robotic Social Attributes scale \cite{carpinella2017robotic}: Warmth, Competence, and Discomfort, and three constructs from the EmCorp-Scale \cite{hoffmann2018peculiarities}: Perception and Interpretation, Tactile Interaction, and Nonverbal Communication. The constructs are continuous values between -3 and 3 and represent the average rating of a 7-point Likert scales across approximately 30 participants for each robot. In particular, we focus on these six constructs because they are averaged across several Likert items. The other constructs reported by the dataset are single Likert items, and thus are not necessarily amenable to regression analysis \cite{schrum2020four}.

\subsection{Creating Features of Robot Embodiment} \label{features}

In this work, we generated three modes of features for each of the robots: hand-crafted (HC) features, metaphor (M) features, and image-based (IM), to be used in the learning process shown in Figure \ref{fig:overview}. Metaphor features and image-based features were deep features that came from large pre-trained models that were available from the transformers library \cite{wolf2020transformers}.  We describe each feature more next.

\subsubsection{Hand-Crafted Features}
Hand-crafted features came entirely from the pre-collected MUFaSAA dataset. These features were characteristics of the robot embodiment that previous research found to be important for human robot interaction (e.g., height \cite{rae2013influence}, waist-to-hip ratio \cite{bernotat2017shape,bernotat2021fe}, presence of a mouth \cite{kalegina2018}, etc.) as well as other features that participants used to describe the robots in the dataset. These features were labeled by annotators that had access to images of the robots and other information from websites created by the robots' manufacturers. The values were all scaled to be between 0 and 1. For each robot, there were 59 HC features.

\begin{table*}
  \begin{threeparttable}
    \caption{Average MSE for Regression Across Different Sets of Features.}
     \begin{tabular}{c|ccc|ccc}
    \hline
        ~ Features Used & Warmth & Competence & Discomfort & Perception and Interpretation & Tactile Interaction & Nonverbal Communication \\ \hline
        
        HC & 0.145** & 0.130* & 0.306* & 0.188* & 0.381*** & 0.182*** \\ 
        M & 0.209 & 0.163 & 0.401 & 0.262 & 1.390 & 0.387 \\ 
        IM & 0.177 & \textbf{0.119}** & 0.344 & 0.202* & 0.445*** & 0.190*** \\ 
        HC + M & 0.137** & 0.134* & 0.311* & 0.184* & 0.388*** & 0.182*** \\ 
        HC + IM & 0.138* & 0.122* & \textbf{0.303}* & \textbf{0.182}* & \textbf{0.337}*** & 0.174*** \\ 
        M + IM & 0.183 & 0.122* & 0.355 & 0.216 & 0.466*** & 0.187*** \\ 
        HC + M + IM & \textbf{0.135}** & 0.124* & 0.307* & \textbf{0.182}* & 0.349*** & \textbf{0.173}*** \\ 
        \hline
        \makecell{Predict Dataset Average \\ (baseline)} & 0.208 & 0.176 & 0.398 & 0.278 & 1.42 & 0.452 \\ \hline
    \end{tabular}
    \begin{tablenotes}
      \item All significance values calculated from a t-test with respect to the baseline's MSE over all folds. * denotes $p<.05$, ** denotes $p<.01$, *** denotes $p<.001$.
    \end{tablenotes}
    \label{tbl:results}
  \end{threeparttable}
\end{table*}

\subsubsection{Metaphor Features}

Metaphor features were created from the three metaphors that were most often used to describe each of the robots in the MUFaSAA dataset. These metaphors consisted of either a single noun  (e.g., ``a dog'', ``a kiosk'', etc.) or the name of a specific reference accompanied by context from where the reference is from (e.g., ``Rosie the Robot from the Jetsons'', ``Eve from WALL-E'', etc.). The metaphors were converted to vectors using a BERT model pre-trained on the Toronto BookCorpus \cite{zhu2015aligning} and English Wikipedia datasets. The pre-trained model output a vector in a 512-dimensional space.

\subsubsection{Image-Based Features}
Image-based features were created from the standardized images of the robots in the MUFaSAA dataset. The images were converted to vectors based on a Vision Transformer (ViT) model that was pre-trained on ImageNet-21k \cite{deng2009imagenet}. The pre-trained model outputs a 512 dimension vector, which we use as our learned features.

\subsection{Regression Experiment}
We formulated understanding user social and functional expectations as a series of regression problems. We used Support Vector Machines (SVMs) to regress robot features onto each of the six constructs in the RoSAS and EmCorp scales. Experiments were conducted in the scikit-learn framework \cite{pedregosa2011scikit}. We selected SVMs because they are often used for datasets of similar size \cite{mathur2020introducing}, and empirically performed the best across all constructs compared to all other regression techniques implemented in scikit-learn. Ground-truth labels came from the user-reported values in the MUFaSAA dataset.

\subsubsection{Model Details}
The SVM regressor used radial basis function as a kernel. The regularization hyperparameter, C, and the margin of tolerance hyperparameter, $\epsilon$, were selected by performing a grid search over the discrete values [.001, .01, .1, 1, 10, 100] for both hyperparameters. These hyperparameters were evaluated by their average mean squared error loss over all constructs and folds in a 20-fold cross-validation setting. The best-performing values were $C=1.0$ and $\epsilon=0.1$.

\subsubsection{Evaluation}
We calculated the average mean squared error (MSE) for each of the six constructs of interest in a 20-fold cross-validation setup, to perform statistical evaluations of our models across folds. We compare our results to the baseline of predicting the average value for the constructs across all robots in the training folds. Statistical analysis is necessary because metrics are noisy on the scale of 165 datapoints, and we seek to verify which combination of modalities performs statistically better than our baseline, which is not captured by point estimates of performance \cite{agarwal2021deep}. We performed this evaluation with every combination of the modalities of describing a robot's embodiment outlined in Section \ref{features}.

\section{Preliminary Results}
The average MSE across the 20-folds for each method are displayed in Table \ref{tbl:results} and show that we can quantitatively predict ratings of social and functional constructs from features of robot embodiments. 

\subsection{Single Mode Results}
We found significant improvements over the baseline with only one mode of feature being used for the HC and IM features. There were no significant differences between the HC and IM features in regressing on the six constructs. This is of particular interest because it suggests that features used from frozen pre-trained networks can be as effective at predicting social and functional expectations as hand-crafted features of robots without the difficulties associated with annotation. 

We did not observe any significant improvements using only metaphor information to predict robot expectations. This may be because the features that can be gained from language models do not contain information on the physical interactions that the metaphors have. Thus, to gain more use from these metaphors, language models may need to ground their understanding of concepts in physical experience \cite{bisk2020experience}. 

\subsection{Multiple Mode Results}
Nearly all combinations of modalities, except M+IM, showed significant improvements over the baseline. In general, the best performing methods involved combinations of multiple modes of features. This suggests that different modes of features have complementary information that is useful in understanding users' mental models. However, these combinations did not show significant improvements over single modes of features.

\section{Future Work and Conclusion}
This work introduces features of robot embodiment to explore how people form mental models of robots. Our results show that features of embodiment can be used to better understand social and functional expectations of robots, and point out several ways to expand this work. In particular, the text-based metaphor features were not as helpful for understanding expectations as the other features. Future work can explore alternate way to calculate these features using other types of information. The MUFaSAA dataset also contains information on frequency of metaphor responses and levels of abstraction that describe how closely the robot resembles each metaphor. This additional information may be useful for generating more informative features for understanding users' social and functional expectations of robots.

We hope that in developing a way to interact with embodiment data, designers in the future may be able to incorporate this data into their design process for both developing new robots and new behaviors for extant robots. By leveraging design information, designers can more accurately understand how their robot will be perceived by general populations. For robot designers, this can be used to decide which features should be included in a robot that either reinforce desired metaphors, or obscure unwanted metaphors. For interaction designers, having this understanding of expectations for embodiment are critical to decide which behaviors are worth developing.

The methods and initial results presented here are preliminary work that shows the potential for features of embodiment to be useful for determining how robots are expected to behave. Notably, there were only static images used to collect the data on social and functional expectations of the robot, however video data may more strongly set these expectations. These results are also subject to other differences based on external factors such as social and  cultural contexts and the experiences of real-world interactions users have with these robots. While the results can be further refined, they show important relationships between how robots are embodied and how they are expected to act, providing insights for the physical and algorithmic design of future robots.

\bibliographystyle{plainnat}
\bibliography{references}

\begin{thebibliography}{31}
\providecommand{\natexlab}[1]{#1}
\providecommand{\url}[1]{\texttt{#1}}
\expandafter\ifx\csname urlstyle\endcsname\relax
  \providecommand{\doi}[1]{doi: #1}\else
  \providecommand{\doi}{doi: \begingroup \urlstyle{rm}\Url}\fi

\bibitem[Agarwal et~al.(2021)Agarwal, Schwarzer, Castro, Courville, and
  Bellemare]{agarwal2021deep}
Rishabh Agarwal, Max Schwarzer, Pablo~Samuel Castro, Aaron~C Courville, and
  Marc Bellemare.
\newblock Deep reinforcement learning at the edge of the statistical precipice.
\newblock \emph{Advances in neural information processing systems},
  34:\penalty0 29304--29320, 2021.

\bibitem[Bernotat et~al.(2017)Bernotat, Eyssel, and Sachse]{bernotat2017shape}
Jasmin Bernotat, Friederike Eyssel, and Janik Sachse.
\newblock Shape it--the influence of robot body shape on gender perception in
  robots.
\newblock In \emph{Social Robotics: 9th International Conference, ICSR 2017,
  Tsukuba, Japan, November 22-24, 2017, Proceedings 9}, pages 75--84. Springer,
  2017.

\bibitem[Bernotat et~al.(2021)Bernotat, Eyssel, and Sachse]{bernotat2021fe}
Jasmin Bernotat, Friederike Eyssel, and Janik Sachse.
\newblock The (fe) male robot: how robot body shape impacts first impressions
  and trust towards robots.
\newblock \emph{International Journal of Social Robotics}, 13:\penalty0
  477--489, 2021.

\bibitem[Bisk et~al.(2020)Bisk, Holtzman, Thomason, Andreas, Bengio, Chai,
  Lapata, Lazaridou, May, Nisnevich, et~al.]{bisk2020experience}
Yonatan Bisk, Ari Holtzman, Jesse Thomason, Jacob Andreas, Yoshua Bengio, Joyce
  Chai, Mirella Lapata, Angeliki Lazaridou, Jonathan May, Aleksandr Nisnevich,
  et~al.
\newblock Experience grounds language.
\newblock \emph{arXiv preprint arXiv:2004.10151}, 2020.

\bibitem[Carpenter et~al.(2009)Carpenter, Davis, Erwin-Stewart, Lee, Bransford,
  and Vye]{carpenter2009gender}
Julie Carpenter, Joan~M Davis, Norah Erwin-Stewart, Tiffany~R Lee, John~D
  Bransford, and Nancy Vye.
\newblock Gender representation and humanoid robots designed for domestic use.
\newblock \emph{International Journal of Social Robotics}, 1\penalty0
  (3):\penalty0 261--265, 2009.

\bibitem[Carpinella et~al.(2017)Carpinella, Wyman, Perez, and
  Stroessner]{carpinella2017robotic}
Colleen~M Carpinella, Alisa~B Wyman, Michael~A Perez, and Steven~J Stroessner.
\newblock The robotic social attributes scale (rosas) development and
  validation.
\newblock In \emph{Proceedings of the 2017 ACM/IEEE International Conference on
  human-robot interaction}, pages 254--262, 2017.

\bibitem[Cha et~al.(2015)Cha, Dragan, and Srinivasa]{cha2015perceived}
Elizabeth Cha, Anca~D Dragan, and Siddhartha~S Srinivasa.
\newblock Perceived robot capability.
\newblock In \emph{2015 24th IEEE International Symposium on Robot and Human
  Interactive Communication (RO-MAN)}, pages 541--548. IEEE, 2015.

\bibitem[Davis(1989)]{davis1989perceived}
Fred~D Davis.
\newblock Perceived usefulness, perceived ease of use, and user acceptance of
  information technology.
\newblock \emph{MIS quarterly}, pages 319--340, 1989.

\bibitem[Deng et~al.(2019)Deng, Mutlu, Mataric, et~al.]{deng2019embodiment}
Eric Deng, Bilge Mutlu, Maja~J Mataric, et~al.
\newblock Embodiment in socially interactive robots.
\newblock \emph{Foundations and Trends{\textregistered} in Robotics},
  7\penalty0 (4):\penalty0 251--356, 2019.

\bibitem[Deng et~al.(2009)Deng, Dong, Socher, Li, Li, and
  Fei-Fei]{deng2009imagenet}
Jia Deng, Wei Dong, Richard Socher, Li-Jia Li, Kai Li, and Li~Fei-Fei.
\newblock Imagenet: A large-scale hierarchical image database.
\newblock In \emph{2009 IEEE conference on computer vision and pattern
  recognition}, pages 248--255. Ieee, 2009.

\bibitem[Dennler et~al.(2023)Dennler, Ruan, Hadiwijoyo, Chen, Nikolaidis, and
  Matari{\'c}]{dennler2023design}
Nathaniel Dennler, Changxiao Ruan, Jessica Hadiwijoyo, Brenna Chen, Stefanos
  Nikolaidis, and Maja Matari{\'c}.
\newblock Design metaphors for understanding user expectations of socially
  interactive robot embodiments.
\newblock \emph{ACM Transactions on Human-Robot Interaction}, 12\penalty0
  (2):\penalty0 1--41, 2023.

\bibitem[Fiske et~al.(2007)Fiske, Cuddy, and Glick]{fiske2007universal}
Susan~T Fiske, Amy~JC Cuddy, and Peter Glick.
\newblock Universal dimensions of social cognition: Warmth and competence.
\newblock \emph{Trends in cognitive sciences}, 11\penalty0 (2):\penalty0
  77--83, 2007.

\bibitem[Hoffmann et~al.(2018)Hoffmann, Bock, and Rosenthal~vd
  P{\"u}tten]{hoffmann2018peculiarities}
Laura Hoffmann, Nikolai Bock, and Astrid~M Rosenthal~vd P{\"u}tten.
\newblock The peculiarities of robot embodiment (emcorp-scale) development,
  validation and initial test of the embodiment and corporeality of artificial
  agents scale.
\newblock In \emph{Proceedings of the 2018 ACM/IEEE international conference on
  human-robot interaction}, pages 370--378, 2018.

\bibitem[Jung et~al.(2017)Jung, Wiltse, Wiberg, and
  Stolterman]{jung2017metaphors}
Heekyoung Jung, Heather Wiltse, Mikael Wiberg, and Erik Stolterman.
\newblock Metaphors, materialities, and affordances: Hybrid morphologies in the
  design of interactive artifacts.
\newblock \emph{Design Studies}, 53:\penalty0 24--46, 2017.

\bibitem[Kalegina et~al.(2018)Kalegina, Schroeder, Allchin, Berlin, and
  Cakmak]{kalegina2018}
Alisa Kalegina, Grace Schroeder, Aidan Allchin, Keara Berlin, and Maya Cakmak.
\newblock \emph{Characterizing the Design Space of Rendered Robot Faces}, page
  96–104.
\newblock Association for Computing Machinery, New York, NY, USA, 2018.
\newblock ISBN 9781450349536.
\newblock URL \url{https://doi.org/10.1145/3171221.3171286}.

\bibitem[Khadpe et~al.(2020)Khadpe, Krishna, Fei-Fei, Hancock, and
  Bernstein]{khadpe2020conceptual}
Pranav Khadpe, Ranjay Krishna, Li~Fei-Fei, Jeffrey Hancock, and Michael
  Bernstein.
\newblock Conceptual metaphors impact perceptions of human-ai collaboration.
\newblock \emph{arXiv preprint arXiv:2008.02311}, 2020.

\bibitem[Kieras and Bovair(1984)]{kieras1984role}
David~E Kieras and Susan Bovair.
\newblock The role of a mental model in learning to operate a device.
\newblock \emph{Cognitive science}, 8\penalty0 (3):\penalty0 255--273, 1984.

\bibitem[Kiesler and Goetz(2002)]{kiesler2002mental}
Sara Kiesler and Jennifer Goetz.
\newblock Mental models of robotic assistants.
\newblock In \emph{CHI'02 extended abstracts on Human Factors in Computing
  Systems}, pages 576--577, 2002.

\bibitem[Kim and Maher(2020)]{kim2020conceptual}
Jingoog Kim and Mary~Lou Maher.
\newblock Conceptual metaphors for designing smart environments: Device, robot,
  and friend.
\newblock \emph{Frontiers in Psychology}, 11:\penalty0 198, 2020.

\bibitem[Kwon et~al.(2016)Kwon, Jung, and Knepper]{kwon2016human}
Minae Kwon, Malte~F Jung, and Ross~A Knepper.
\newblock Human expectations of social robots.
\newblock In \emph{2016 11th ACM/IEEE International Conference on Human-Robot
  Interaction (HRI)}, pages 463--464. IEEE, 2016.

\bibitem[Mathur and Matari{\'c}(2020)]{mathur2020introducing}
Leena Mathur and Maja~J Matari{\'c}.
\newblock Introducing representations of facial affect in automated multimodal
  deception detection.
\newblock In \emph{Proceedings of the 2020 International Conference on
  Multimodal Interaction}, pages 305--314, 2020.

\bibitem[Moon and Nass(1996)]{moon1996real}
Youngme Moon and Clifford Nass.
\newblock How “real” are computer personalities? psychological responses to
  personality types in human-computer interaction.
\newblock \emph{Communication research}, 23\penalty0 (6):\penalty0 651--674,
  1996.

\bibitem[Pedregosa et~al.(2011)Pedregosa, Varoquaux, Gramfort, Michel, Thirion,
  Grisel, Blondel, Prettenhofer, Weiss, Dubourg, et~al.]{pedregosa2011scikit}
Fabian Pedregosa, Ga{\"e}l Varoquaux, Alexandre Gramfort, Vincent Michel,
  Bertrand Thirion, Olivier Grisel, Mathieu Blondel, Peter Prettenhofer, Ron
  Weiss, Vincent Dubourg, et~al.
\newblock Scikit-learn: Machine learning in python.
\newblock \emph{the Journal of machine Learning research}, 12:\penalty0
  2825--2830, 2011.

\bibitem[Perugia et~al.(2022)Perugia, Guidi, Bicchi, and
  Parlangeli]{perugia2022shape}
Giulia Perugia, Stefano Guidi, Margherita Bicchi, and Oronzo Parlangeli.
\newblock The shape of our bias: Perceived age and gender in the humanoid
  robots of the abot database.
\newblock In \emph{2022 17th ACM/IEEE International Conference on Human-Robot
  Interaction (HRI)}, pages 110--119. IEEE, 2022.

\bibitem[Phillips et~al.(2018)Phillips, Zhao, Ullman, and
  Malle]{phillips2018human}
Elizabeth Phillips, Xuan Zhao, Daniel Ullman, and Bertram~F Malle.
\newblock What is human-like? decomposing robots' human-like appearance using
  the anthropomorphic robot (abot) database.
\newblock In \emph{Proceedings of the 2018 ACM/IEEE international conference on
  human-robot interaction}, pages 105--113, 2018.

\bibitem[Rae et~al.(2013)Rae, Takayama, and Mutlu]{rae2013influence}
Irene Rae, Leila Takayama, and Bilge Mutlu.
\newblock The influence of height in robot-mediated communication.
\newblock In \emph{2013 8th ACM/IEEE International Conference on Human-Robot
  Interaction (HRI)}, pages 1--8. IEEE, 2013.

\bibitem[Rueben et~al.(2021)Rueben, Klow, Duer, Zimmerman, Piacentini,
  Browning, Bernieri, Grimm, and Smart]{rueben2021mental}
Matthew Rueben, Jeffrey Klow, Madelyn Duer, Eric Zimmerman, Jennifer
  Piacentini, Madison Browning, Frank~J Bernieri, Cindy~M Grimm, and William~D
  Smart.
\newblock Mental models of a mobile shoe rack: exploratory findings from a
  long-term in-the-wild study.
\newblock \emph{ACM Transactions on Human-Robot Interaction (THRI)},
  10\penalty0 (2):\penalty0 1--36, 2021.

\bibitem[Schrum et~al.(2020)Schrum, Johnson, Ghuy, and
  Gombolay]{schrum2020four}
Mariah~L Schrum, Michael Johnson, Muyleng Ghuy, and Matthew~C Gombolay.
\newblock Four years in review: Statistical practices of likert scales in
  human-robot interaction studies.
\newblock In \emph{Companion of the 2020 ACM/IEEE International Conference on
  Human-Robot Interaction}, pages 43--52, 2020.

\bibitem[Voida et~al.(2008)Voida, Mynatt, and Edwards]{voida2008re}
Stephen Voida, Elizabeth~D Mynatt, and W~Keith Edwards.
\newblock Re-framing the desktop interface around the activities of knowledge
  work.
\newblock In \emph{Proceedings of the 21st annual ACM symposium on User
  interface software and technology}, pages 211--220, 2008.

\bibitem[Wolf et~al.(2020)Wolf, Debut, Sanh, Chaumond, Delangue, Moi, Cistac,
  Rault, Louf, Funtowicz, et~al.]{wolf2020transformers}
Thomas Wolf, Lysandre Debut, Victor Sanh, Julien Chaumond, Clement Delangue,
  Anthony Moi, Pierric Cistac, Tim Rault, R{\'e}mi Louf, Morgan Funtowicz,
  et~al.
\newblock Transformers: State-of-the-art natural language processing.
\newblock In \emph{Proceedings of the 2020 conference on empirical methods in
  natural language processing: system demonstrations}, pages 38--45, 2020.

\bibitem[Zhu et~al.(2015)Zhu, Kiros, Zemel, Salakhutdinov, Urtasun, Torralba,
  and Fidler]{zhu2015aligning}
Yukun Zhu, Ryan Kiros, Rich Zemel, Ruslan Salakhutdinov, Raquel Urtasun,
  Antonio Torralba, and Sanja Fidler.
\newblock Aligning books and movies: Towards story-like visual explanations by
  watching movies and reading books.
\newblock In \emph{Proceedings of the IEEE international conference on computer
  vision}, pages 19--27, 2015.

\end{thebibliography}

\end{document}